\documentclass[journal]{IEEEtran}

\usepackage{hyperref}
\usepackage{cite}
\usepackage[pdftex]{graphicx}
\usepackage{amsmath, amssymb}
\usepackage[caption=false,font=footnotesize]{subfig}
\usepackage{rotating}

\graphicspath{{figures/}}

\begin{document}
\title{Hierarchical Mixtures of Generators for Adversarial Learning}
\author{Alper~Ahmeto\u{g}lu,~Ethem~Alpayd{\i}n~\IEEEmembership{}
\thanks{A. Ahmeto\u{g}lu is with the Department of Computer Engineering, Bo{\u g}azi{\c c}i University, Bebek Istanbul 34342 Turkey, e-mail: ahmetoglu.alper@gmail.com.}
\thanks{E. Alpayd{\i}n is with the Department of Computer Science, {\"O}zye{\u g}in University, {\c C}ekmek{\"o}y Istanbul 34794 Turkey, e-mail: ethem.alpaydin@ozyegin.edu.tr.}
}
\maketitle

\begin{abstract}
Generative adversarial networks (GANs) are deep neural networks that allow us to sample from an arbitrary probability distribution without explicitly estimating the distribution. There is a generator that takes a latent vector as input and transforms it into a valid sample from the distribution. There is also a discriminator that is trained to discriminate such fake samples from true samples of the distribution; at the same time, the generator is trained to generate fakes that the discriminator cannot tell apart from the true samples. Instead of learning a global generator, a recent approach involves training multiple generators each responsible from one part of the distribution. In this work, we review such approaches and propose the hierarchical mixture of generators, inspired from the hierarchical mixture of experts model, that learns a tree structure implementing a hierarchical clustering with soft splits in the decision nodes and local generators in the leaves. Since the generators are combined softly, the whole model is continuous and can be trained using gradient-based optimization, just like the original GAN model. Our experiments on five image data sets, namely, MNIST, FashionMNIST, UTZap50K, Oxford Flowers, and CelebA, show that our proposed model generates samples of high quality and diversity in terms of popular GAN evaluation metrics. The learned hierarchical structure also leads to knowledge extraction.
\end{abstract}


\IEEEpeerreviewmaketitle

\def\cX{{\cal X}}
\section{Introduction}
\label{sec:introduction}

In generative modeling, we are given a data set $\cX=\{x^t\}_t$ sampled from some unknown probability distribution $p(x)$ and we want to be able to generate new instances from $p(x)$. This is an unsupervised learning problem and the usual approach is to first build an estimator for $p(x)$ and then sample from that. The generative adversarial network (GAN) \cite{goodfellow2014generative} is interesting in that it learns a generative model without explicitly modeling $p(x)$ but by using an auxiliary discriminative model, thereby transforming an unsupervised learning problem into a supervised learning problem. 

A GAN model is composed of two learners, a generator $G$ and a discriminator $D$. $G$ takes as input random $z$ drawn from some simple parametric distribution of relatively low dimensionality, e.g., a zero-mean Gaussian with unit covariance, and learns to transform it to a valid instance $\tilde{x}$ from (the unknown) $p(x)$. $G$ is implemented as a deep neural network that takes $z$ as input, generates $\tilde{x}$ as output, and has many layers in between necessary for the transformation; the weights in $G$ are denoted by $\theta$. The $\tilde{x}$ that are generated by $G$ are called {\em fake\/} because they are synthetic. The discriminator $D$ is a two-class classifier that learns to discriminate such fakes from true $x^t$ sampled from the training set $\cX$. $D$ is another deep neural network with either $\tilde{x}$ or $x^t$ as input and 0 or 1 as the desired output respectively for the single sigmoid output. Again $D$ has as many hidden layers as necessary for the task; the weights in $D$ are denoted by $\phi$.

The objective function is
\begin{equation}
\label{eq:gan}
\min_{\theta} \max_{\phi} \mathbb{E}_{x^t \sim p(x)} [ \log{D(x;\phi)} ] + 
\mathbb{E}_{z \sim p(z)} [ \log{(1-D(G(z;\theta);\phi))} ]
\end{equation}

We train the weights of both $G$ and $D$ using gradient-based optimization, alternating between the two. $D$ wants to maximize the likelihood for true instances $x^t$ (drawn from unknown $p(x)$ as represented by the training set $\cX$) and minimize the likelihood for fake instances generated by $G$. At the same time, $G$ wants to generate fakes for which $D$ assigns as high likelihoods as possible. As $G$ gets better in generating fakes for which $D$ assigns high likelihood, $D$ is forced to better separate them from true instances, which in turn forces $G$ to generate even better fakes, and so on.

GANs are used successfully especially in image generation. A well-trained GAN can generate images that are almost indistinguishable by humans \cite{karras2017progressive,brock2018large,karras2019style}; still, there are two main difficulties in training: Sometimes $G$ learns only a part of the true $p(x)$ and can  generate only a subset of the possible $x$; this is called 
{\em mode collapse\/} because it is indication that $G$ does not cover all the modes of $p(x)$. The second problem is that of {\em vanishing gradients\/} that we always have in training deep neural networks; note that here because both $D$ and $G$ are deep, $G$ is doubly deep because its gradient needs to be back-propagated through $D$.

There is recent work in the literature that focuses on these problems. To solve problems related to training, it has been proposed to use either different objective functions, regularization methods, or architectures; see \cite{creswell2018generative,hong2019generative,kurach2018gan} for good surveys of the state of the art.

The direction we pursue in this study is to use multiple generators each one responsible from generating a local region of $p(x)$. Different local generators will learn to cover different modes and this will help alleviate the mode collapse problem. They also help with the problem of vanishing gradients because local generators are simpler, i.e., more shallow, and hence the paths through which the gradient is back-propagated are shorter.
We review three previously proposed approaches from the literature, namely multi-agent diverse GAN (MADGAN)\cite{ghosh2018multi}, mixture GAN (MGAN)\cite{hoang2018mgan}, and mixtures of experts GAN (MEGAN)\cite{park2018megan}.

We propose the hierarchical mixture of generators that has a tree structure with internal decision nodes that divide up the latent $z$ space and leaves that are local generators responsible from a local $z$ region generating a subset of $p(x)$. Since the splits are soft, given the tree structure, the split parameters at the internal nodes as well as those of the generators in the leaves can be updated using gradient-descent. Note that it is only $G$ that is modeled this way and split locally, there is still a single $D$ implemented as a deep fully-connected neural network as usual.

The rest of this paper is organized as follows. In Chapter \ref{sec:combining}, we discuss the previously proposed models in literature that also use multiple generators. We explain our proposed model of hierarchical mixture of generators in detail in Chapter \ref{sec:mixture}. Our experimental results on a toy two-dimensional and five real-world image data sets are given in Chapter \ref{sec:experiments}. We conclude in Chapter \ref{sec:conclusion}.

\section{Combining multiple generators in GAN}
\label{sec:combining}

\subsection{Multi-agent diverse GAN}
\label{subsec:madgan}
In the multi-agent diverse GAN (MADGAN) \cite{ghosh2018multi}, there are $K\ge 2$ generators each of which labels the fake data it generates with its index. $D$ does not learn a two-class, true vs. fake classification problem, but a $K+1$-class problem where class 0 is for true instances and classes 1 to $K$ are the different ways of generating fake; in terms of implementation, $D$ has $K$ softmax outputs instead of one sigmoid output as we have for the original GAN. 

\begin{figure}[htbp]
\begin{center}
\includegraphics[width=0.95\columnwidth]{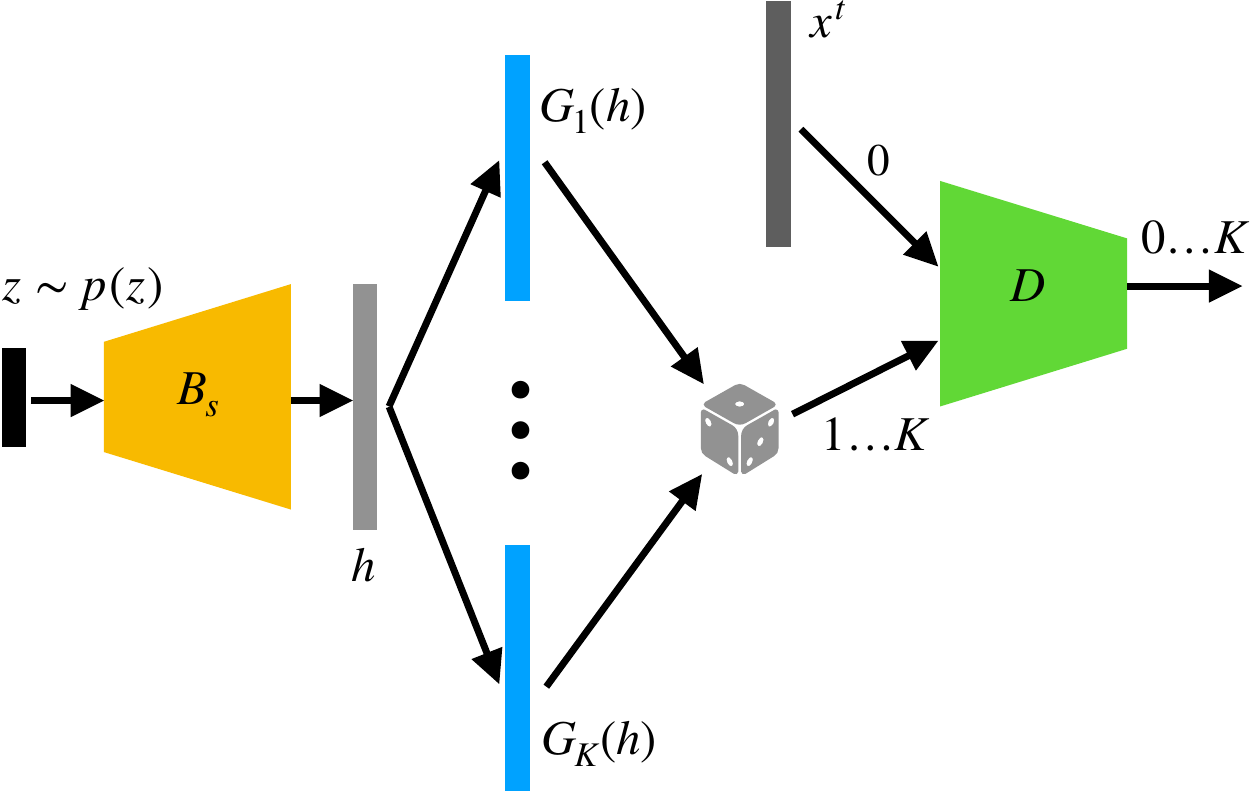}
\end{center}
\caption{Multi-agent diverse GAN (MADGAN)\cite{ghosh2018multi}. Given the latent $z$, the shared network $B_s$ (shown in yellow) produces an intermediate representation $h$. From this representation, each generator $G_i$ (shown in blue) outputs a sample, one of which is randomly chosen and fed to the discriminator $D$ as a fake together with the index of its generator. $D$ (shown in green) is a $K+1$-class classifier.}
\label{fig:models:madgan}
\end{figure}

The model is shown in Figure \ref{fig:models:madgan}. Given $z$, first a shared neural network block $B_s$ produces an intermediate representation $h$. $z$ is a low-dimensional unstructured vector, $B_s$ contains successive deconvolution layers and generates a two-dimensional image using multiple filters in $h$, which is given as input to a set of generators $G_i, i=1,\ldots,K$, one of which we choose at random. The discriminator sees either a true $x^t$ with class code 0 or one of the generated fake with its index as the class code. The discriminator should push the different generators to different modes to solve the classification problem successfully: For correct classification, instances from the same class (i.e., fakes generated by the same generator) need to be more similar than instances from different classes (i.e., fakes generated by different generators).

More formally, for the discriminator, the objective is to 
\begin{equation}
\max_{\phi} \mathbb{E}_{x^t \sim p(x)} \left[ \log D_0(x^t;\phi)\right] + \mathbb{E}_{z^t \sim p(z)} \left[ \sum_{i=1}^K r^t_i \log D_i(G_i(z;\theta);\phi)\right]
\label{eq:madgan1}
\end{equation}

\noindent where $r^t_i$ is 1 if $z^t$ is processed by generator $i$ and 0 otherwise, and $D_i$ denotes the output of the discriminator for class $i=0,\ldots,K$. 

In updating the weights of $G_i$, the objective is to 
\begin{equation}
\min_{\theta} \mathbb{E}_{z^t \sim p(z)} \left[\log (1-D_0(G_i(z^t;\theta))) \right]
\label{eq:madgan2}
\end{equation}

Note that though there are multiple generators, their outputs are not combined in a cooperative manner. We do not partition $p(z)$ and use each local partition for a different generators; for any $z$, any of the generators can be used. It is more as if each generator produces its own interpretation of $p(z)$; instead of partitioning $p(z)$, we learn alternative generator functions for the same region in $p(z)$.

\subsection{Mixture GAN}
\label{subsec:mgan}

The mixture GAN (MGAN) \cite{hoang2018mgan} has some similarities with MADGAN, the main difference being that the classifier and the discriminator are separated. The discriminator is a two-class classifier as usual differentiating between true and fake examples, and there is an additional $K$-class classifier used only for the fake examples learning the index of the generator used.

\begin{figure}[htbp]
\begin{center}
\includegraphics[width=0.95\columnwidth]{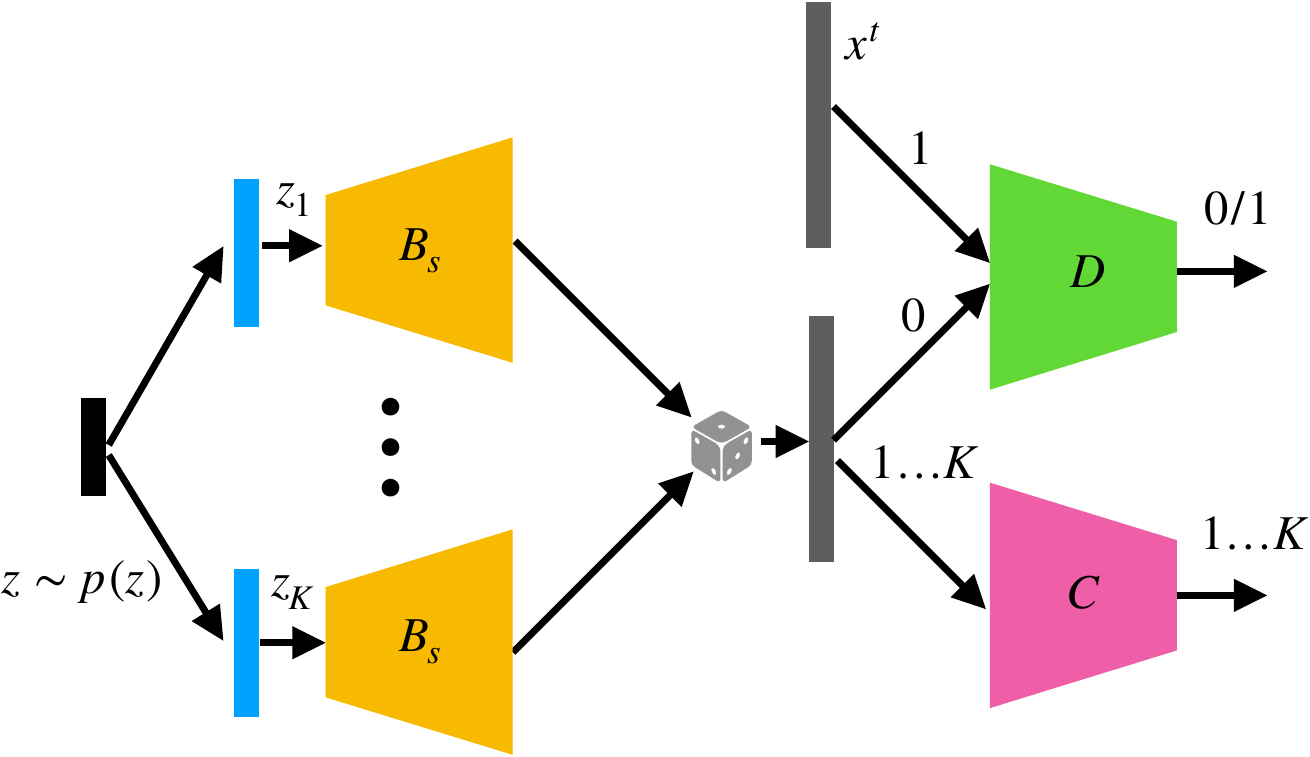}
\end{center}
\caption{Mixture GAN \cite{hoang2018mgan}. This is similar to MADGAN architecture with one difference being that the multiple generators are used earlier: Each generator $G_i$ creates an abstract representation $z_i$ which is fed to the shared block $B_s$ to generate the output. One of the $K$ generated fakes is selected at random and given to the discriminator; there is also an additional classifier $C$ (shown in purple) that learns the index of the generator of a fake.}
\label{fig:models:mgan}
\end{figure}

The model is shown in Figure \ref{fig:models:mgan}. There is also the difference that the split of the generators is earlier and the shared deconvolution block $B_s$ comes afterwards. The generators $G_i, i=1,\ldots,K$ transform $z$ to $z_i$ in parallel and for all, the shared $B_s$ produces the final output. Training is formalized as a multi-task learning problem where the discriminator is trained to discriminate between the fake and the real data as usual, and at the same time, for a fake, the $K$-class classifier tries to predict the index of the generator that produced it. 

The overall objective is defined as follows:
\begin{equation}
\begin{split}
\min_{\theta,\psi} \max_{\phi} &\ \mathbb{E}_{x^t \sim p(x)} \left[ \log D(x^t;\phi) \right]\\ &+ \mathbb{E}_{z^t \sim p(z)} \left[\sum_{i=1}^K r_i^t \log (1-D(G_i(z^t;\theta);\phi))\right]\\
&-\mathbb{E}_{z^t \sim p(z)} \left[\sum_{i=1}^K r_i^t\log (C_i(G_i(z^t;\theta);\psi))\right]
\end{split}
\end{equation}

\noindent where $C$, parameterized by $\psi$, is the $K$-class classifier for the fakes whose output for class $i$ is denoted by $C_i$.

\subsection{Mixtures of experts GAN}
\label{subsec:megan}

In MEGAN \cite{park2018megan}, inspired from the mixtures of experts architecture \cite{jacobs1991adaptive}, there is an additional gating model, which is also trained, that chooses among the different generators.

\begin{figure}[htbp]
\begin{center}
\includegraphics[width=0.9\columnwidth]{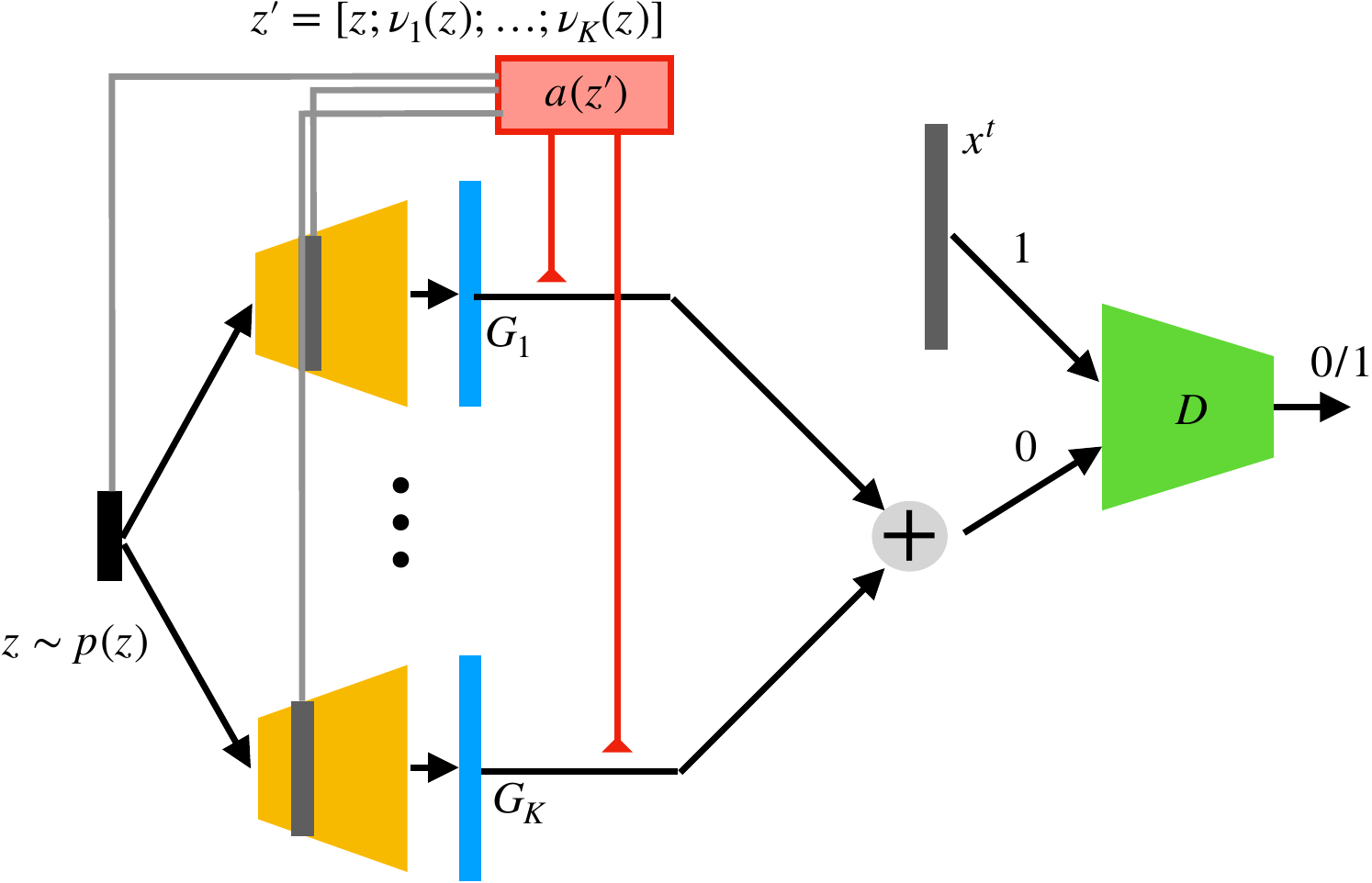}
\end{center}
\caption[Mixture of experts GAN]{Mixture of experts GAN \cite{park2018megan}. Given $z$, each generator $G_i$ outputs a sample $x_i$. The gating network (shown in red) selects one generated sample among all based on $z$ and $\nu_i(z)$, which are the first layer activations of the generators. Unlike the original mixtures of experts \cite{jacobs1991adaptive}, the selection is hard; only one generator is used.}
\label{fig:models:megan}
\end{figure}

The model is shown in Figure \ref{fig:models:megan}. In addition to the generators $G_i, i=1,\ldots,K$, there is also a gating network that takes $z$ and $\nu_i(z), i=1,\ldots,K$ as its input: $\nu_i(z)$ are the first layer activations of $G_i$ and as such are believed to provide additional information as to how best to choose the responsible generator. Then Straight-Through Gumbel Softmax is applied which only selects one expert while allowing differentiability. The discriminator is still two-class. The gating model also has its parameters that are updated together with the generators. Although all generators generate an output, it is the gating model that decides which one is to be used. Except the way the generator is written as a weighted sum of generators, the training objective is the same as Equation (\ref{eq:gan}) used in the original GAN model.

Different from MADGAN and MGAN, here, $p(z)$ is partitioned into local regions which map to local regions of $p(x)$. Thanks to the gating network outputs, each generator $G_i$ is only responsible for a local region of $p(z)$ and generates the corresponding local region of $p(x)$. However, this partitioning is hard since we let only one generator to be used. Besides, the gating network takes processed features $\nu_i(z)$ as extra inputs and this may lead to a partitioning that might be non-smooth in the $z$ space.

\section{Hierarchical mixtures of generators}
\label{sec:mixture}

All previous approaches use multiple generators, yet these generators do not work cooperatively, and they all train a flat set of generators. We propose the hierarchical mixture of generators that are inspired from the hierarchical mixtures of experts \cite{jordan1994hierarchical}, where the generators are organized at the leaves of a tree and they cooperate as defined by the tree structure.

Let us think of a binary decision tree. The generators are at the leaves of this tree. At each internal node $m$ of the tree, there is a gating function $\sigma_m(z)$ with parameters $\{v_m, v_{m_0}\}$ that calculates the probability that we take the left child
\begin{equation}
\sigma_m(z) = \frac{1}{1+\exp{[-(v_m z + v_{m0})]}}
\label{eq:sigmoid}
\end{equation}

\begin{figure}[htbp]
\begin{center}
\includegraphics[width=0.98\columnwidth]{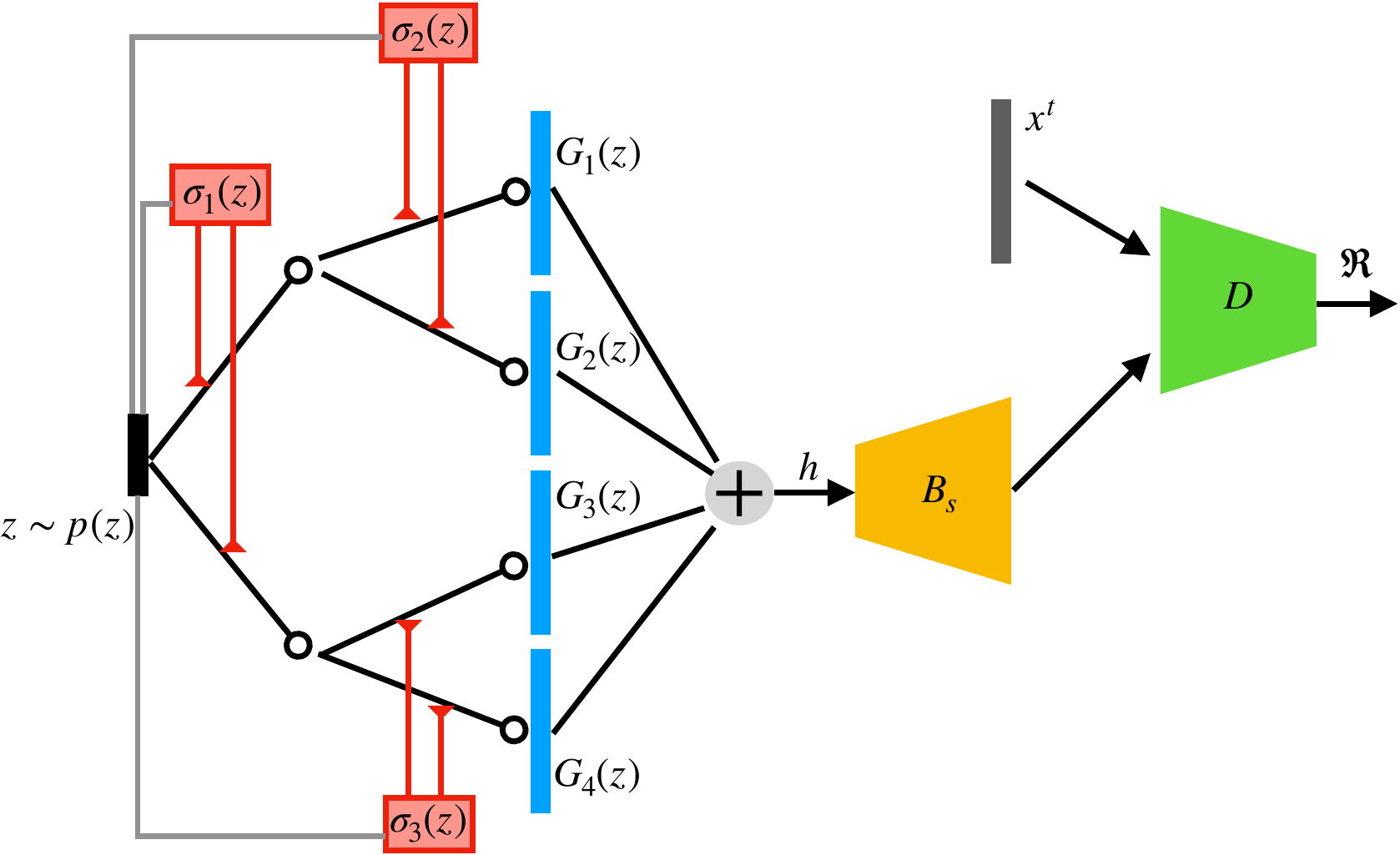}
\end{center}
\caption{Hierarchical mixture of generators (HMoG) with depth two and four generators. Each generator creates an intermediate representation from $z$ and their average is calculated with the weights given by gating values on each path. The resulting $h$ is passed through $B_s$ to generate the image.}
\label{fig:models:hmog}
\end{figure}

$1-\sigma_m(z)$ is the probability that we take the right child. If $m$ is a leaf mode, the response is given by the generator at that leaf, $G_m(z)$. If $m$ is an internal node, its response is a weighted sum of its left and right children weighted by the gating values:
\begin{equation}
x_m(z)=
	\begin{cases}
		\hfil G_m(z) &\text{if $m$ is a leaf} \\
		\hfil x_m^{L}(z) \sigma_m(z) + x_m^{R}(z)(1 - \sigma_m(z)) &\text{otherwise}
	\end{cases}
\label{eq:hmgan2}
\end{equation}
where $x_m^L$ and $x_m^R$ are the responses of the left and the right children respectively, calculated recursively until we get to leaf nodes; see Figure \ref{fig:models:hmog}. 

The generators at the leaves are simple linear models:
\begin{equation}
G_m(z) = W_m z + w_{m0} \label{eq:linear}
\end{equation}

Because the gating in Equation (\ref{eq:sigmoid}) is a sigmoid, we take a soft combination of generators at the leaves. This has two uses: First, we have a smooth transition from one local generator to another smoothly interpolating in between. Second, the model is differentiable and therefore given a tree structure, we can use gradient-based optimization to learn all the gating parameters $\{v_m,v_{m0}\}$ in the decision nodes and the parameters of the generators $\{W_m,w_{m0}\}$ at the leaves.

In training, we use the Wasserstein loss \cite{arjovsky2017wasserstein} which has been shown to work better than the likelihood-based criterion of Equation (\ref{eq:gan}):
\begin{equation}
	\min_{\theta} \max_{\phi} \quad \mathbb{E}_{x^t \sim p(x)} [ D(x^t; \phi) ] - \mathbb{E}_{z^t \sim p(z)} [ D(G(z^t;\theta);\phi) ]
\end{equation}

Here, $D$ estimates the score of ``trueness'' for a sample, and Wasserstein loss checks for the difference between the average scores for true samples and the generated fake samples. $D$, which is a regressor and not a classifier, is trained to maximize this, and $G$ is trained to minimize it.

\begin{figure}[htbp]
\begin{center}
\includegraphics[width=0.98\columnwidth]{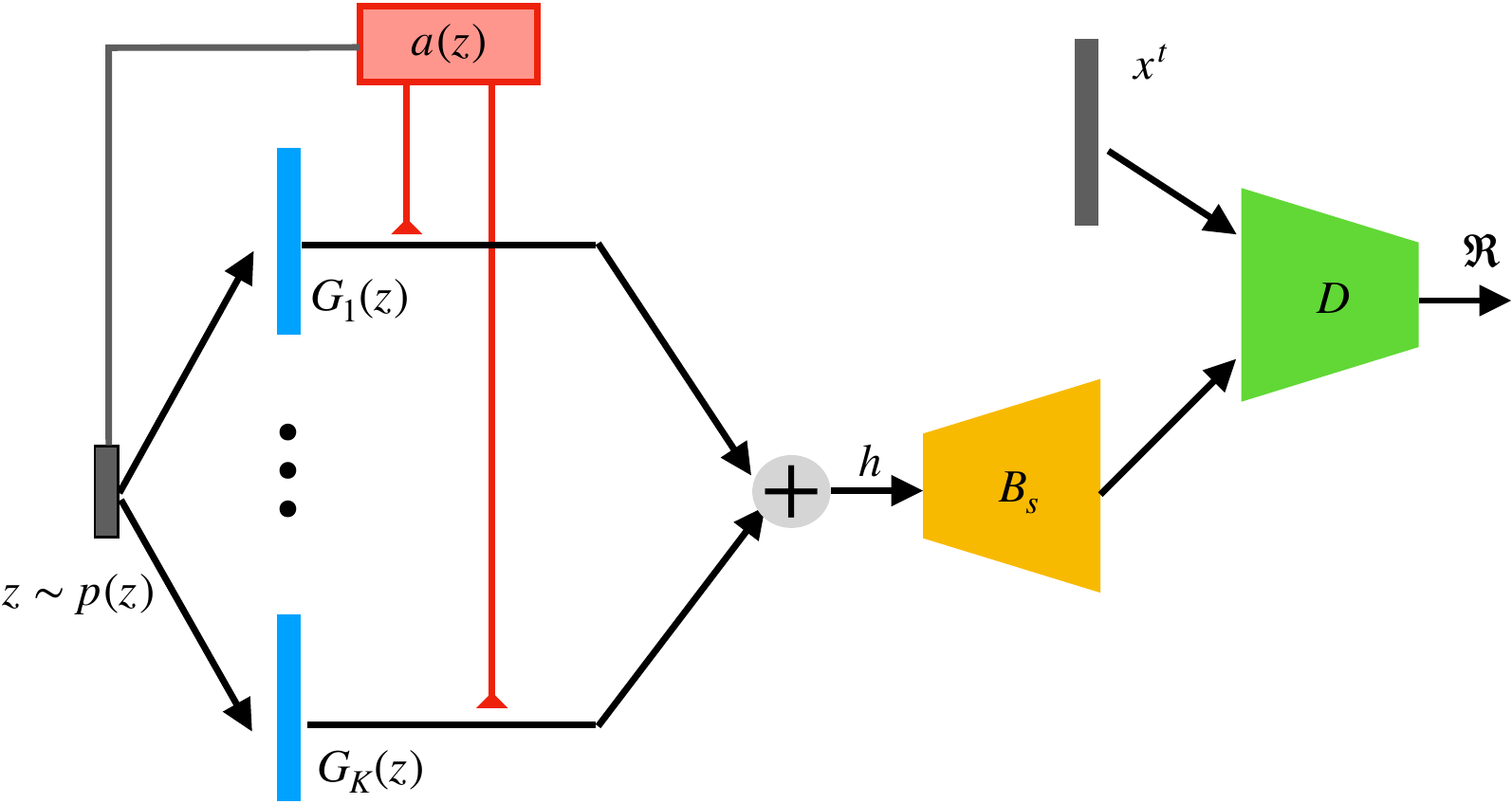}
\end{center}
\caption{Mixture of generators (MoG). Given $z$, each generator outputs an intermediate $h$. The gating unit chooses among $K$ generators using the softmax that assigns weights that are between 0 and 1 and sum up to 1. Hence, the output is a convex combination of all the generators, which is then passed through $B_s$ to generate the image.}
\label{fig:models:mog}
\end{figure}

Note that Equation (\ref{eq:sigmoid}) defines a binary tree; we can also have a $K>2$-ary tree by using the softmax instead of the sigmoid in the gating nodes. At the extreme, as a special case, we can have a tree of depth 1 with $K$ generator leaves; see Figure \ref{fig:models:mog}. This is the (flat) mixtures of generators (MoG), which is similar to MEGAN with two differences: We keep softmax gating so the combination is soft just like the original mixture of experts model \cite{jacobs1991adaptive}, and the input to gating uses only $z$ without any extra features $\nu_i(z)$ extracted from $G_i$.

\section{Experiments}
\label{sec:experiments}

\subsection{Results on toy data}
\label{subsec:results_toy}

\begin{figure*}[htbp]
\begin{center}
\subfloat[MADGAN]{
    \includegraphics[width=0.3\textwidth]{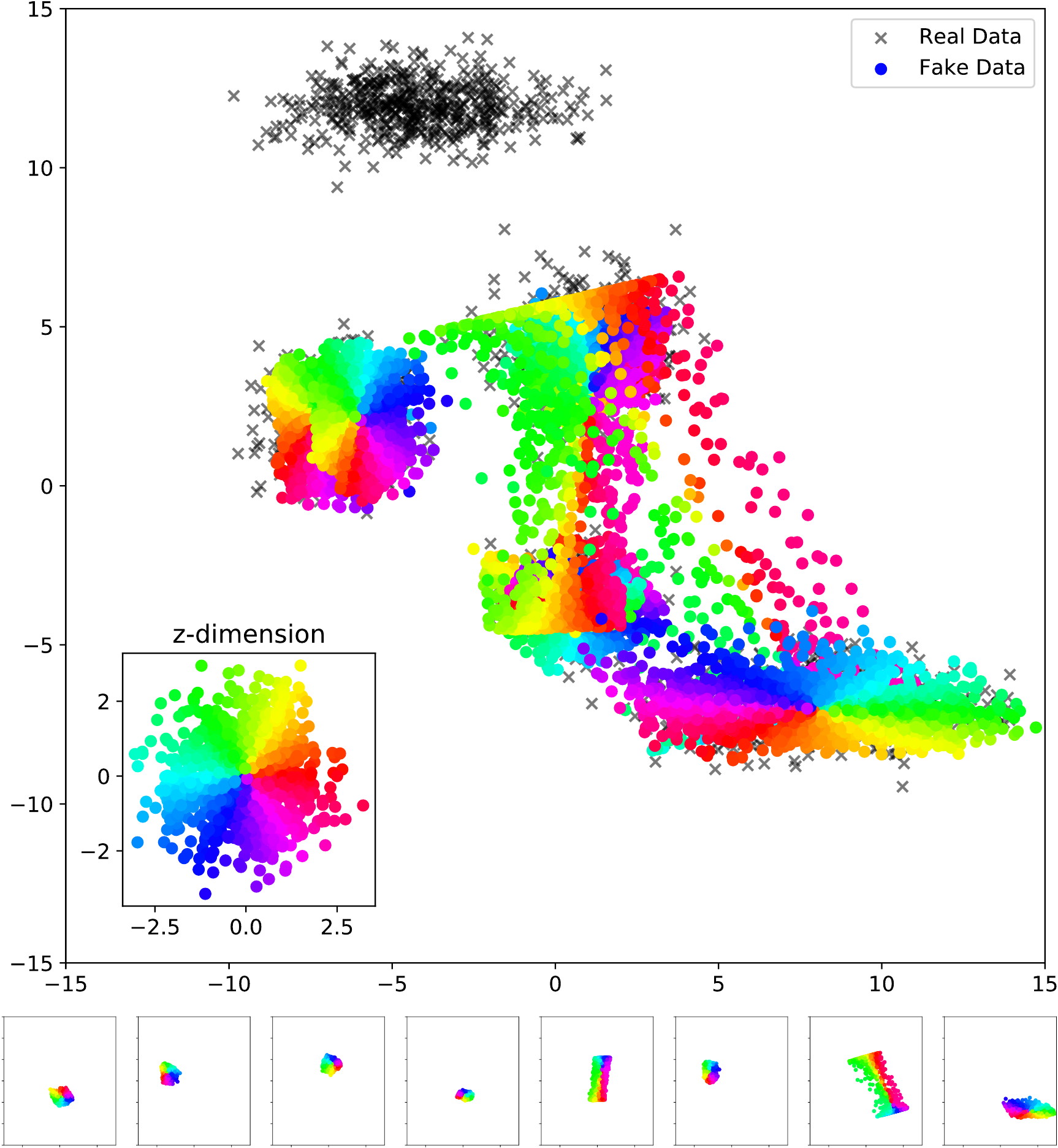}
}
\subfloat[MGAN]{
    \includegraphics[width=0.3\textwidth]{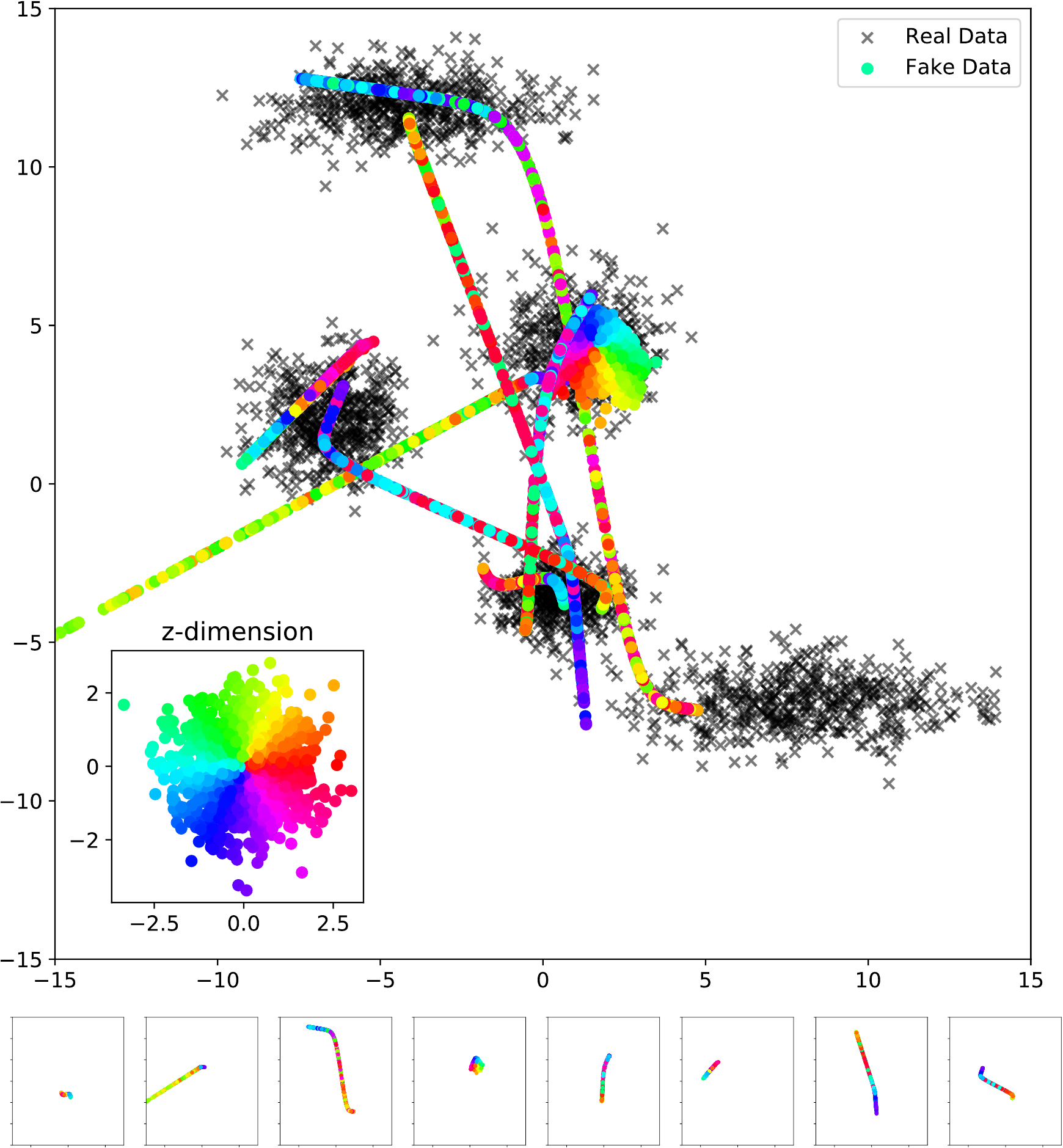}
}
\subfloat[MEGAN]{
    \includegraphics[width=0.3\textwidth]{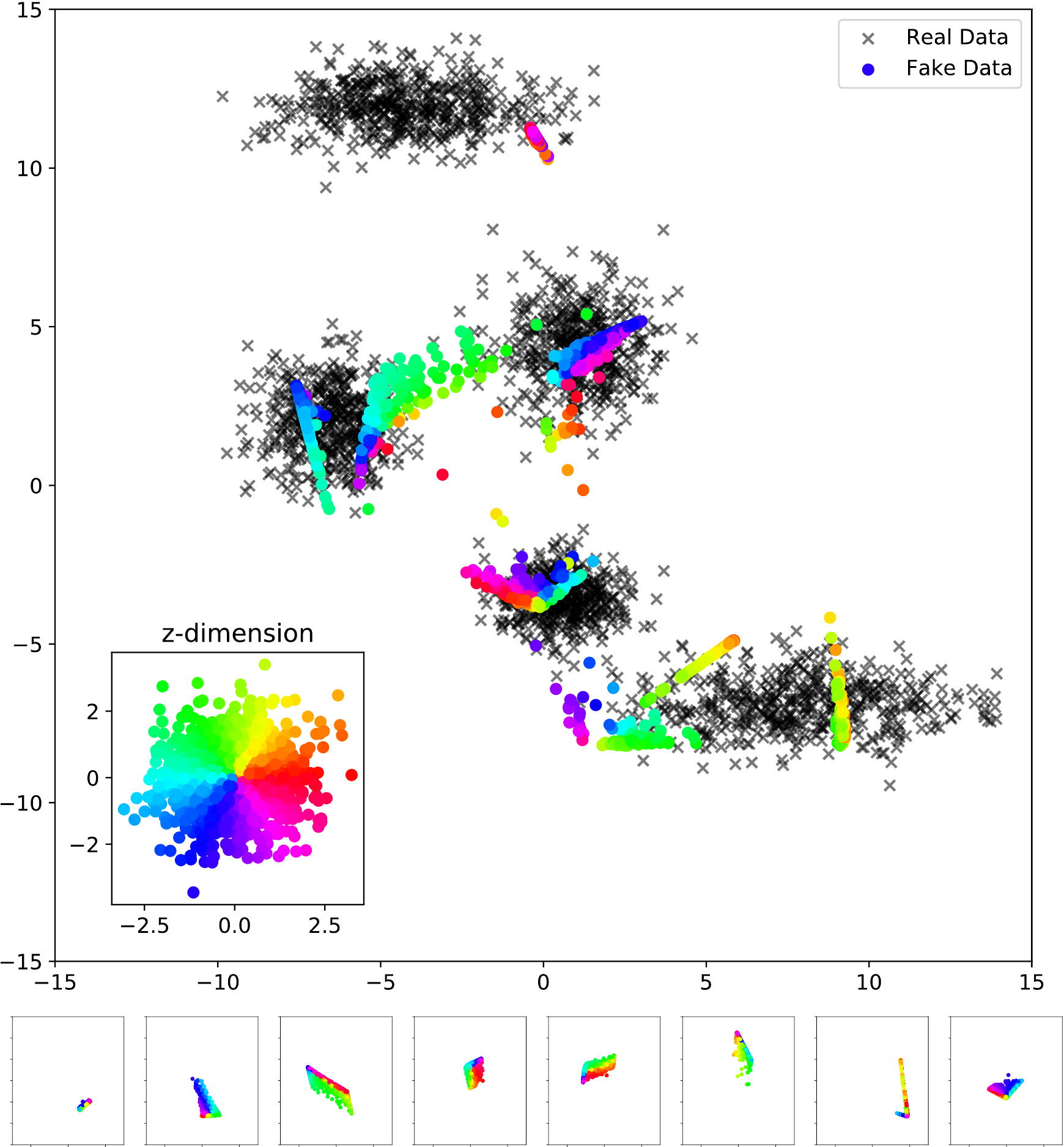}
}
\end{center}
\caption{Results using the three competing methods that also use multiple generators. In all three, in the larger figure above, the toy data set is shown with black crosses and the different colors represent different parts of $p(z)$ which is similarly color coded in the lower left hand side. The small figures below show the part of data generated by the eight individual generators.}
\label{fig:toy_others}
\end{figure*}

\begin{figure}[htbp]
\begin{center}
\includegraphics[width=0.95\columnwidth]{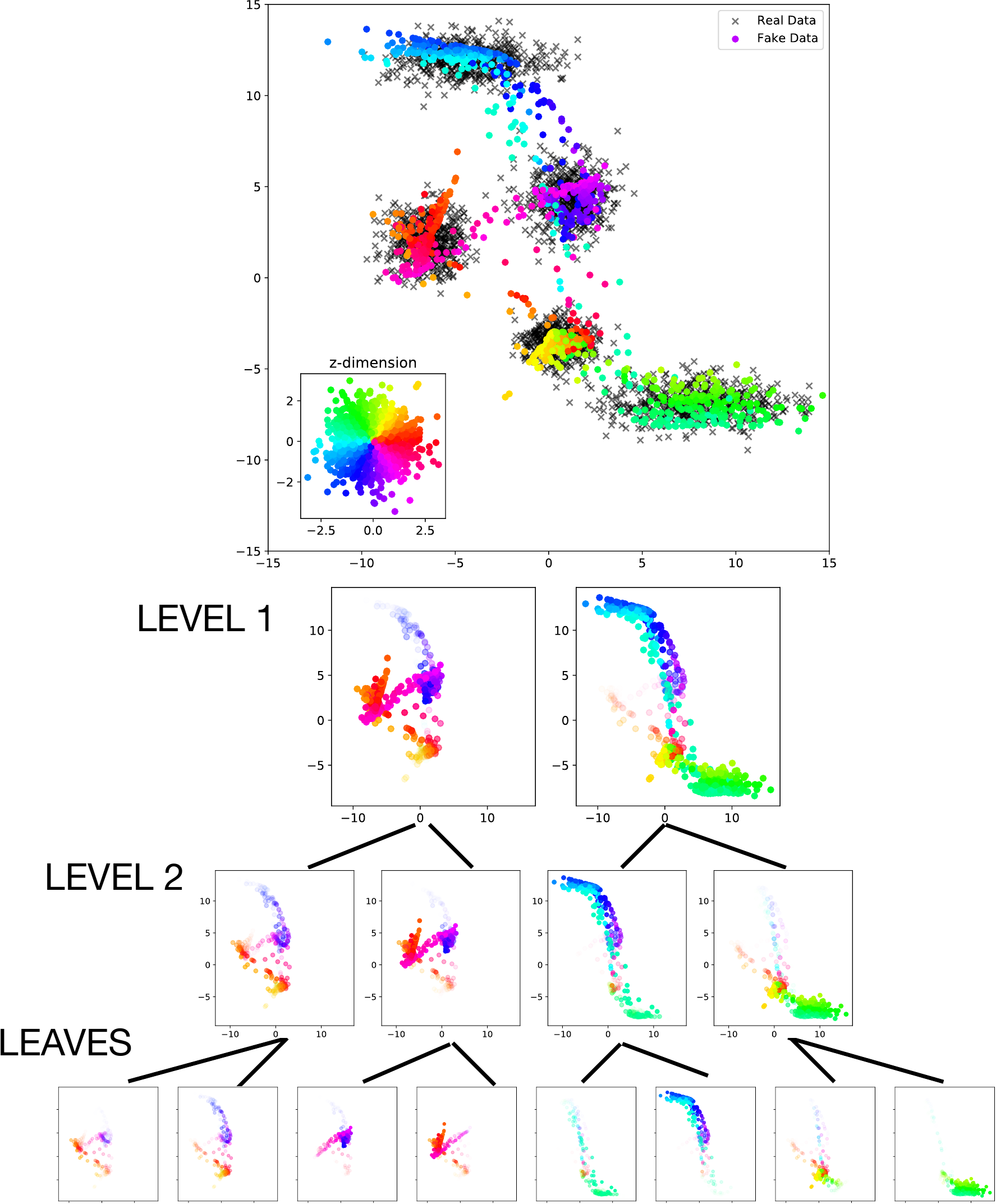}
\end{center}
\caption{Result using our proposed hierarchical mixture of generators (HMoG) with depth three and eight generators on the toy data. How the data is split over the tree at various levels and the responsibilities of the generators at the leaves are also shown.}
\label{fig:toy_hmog}
\end{figure}

\begin{figure}[htbp]
\begin{center}
\includegraphics[width=0.5\columnwidth]{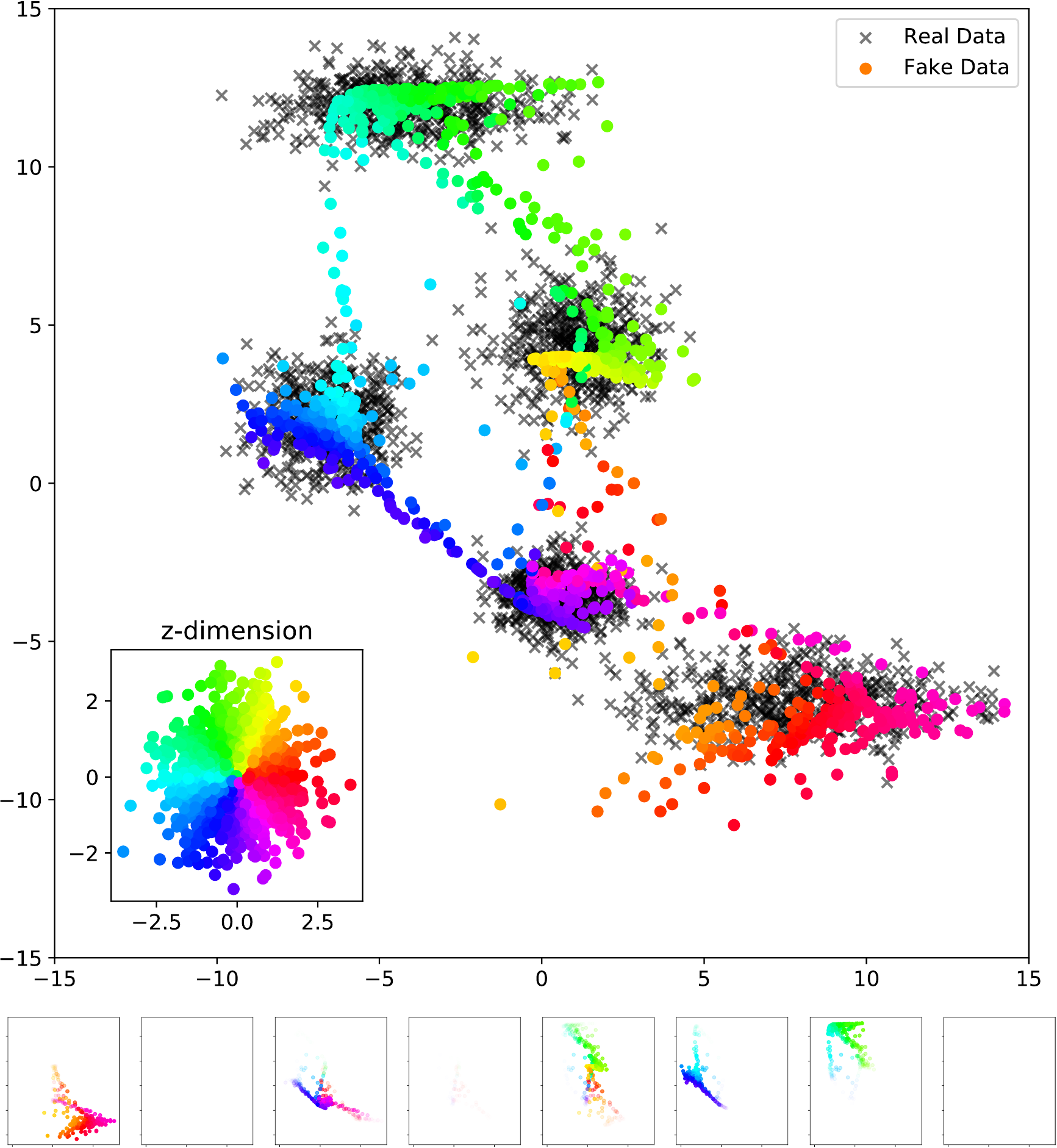}
\end{center}
\caption{Result using the flat mixture of generators (MoG) on the toy data. The small figures below show the part of data generated by each generator.}
\label{fig:toy_mog}
\end{figure}

We begin with experiments on a toy two-dimensional data set sampled from a mixture of five Gaussians. The latent $z$ is drawn from a two-dimensional Gaussian distribution with zero mean and unit variance, and we use eight generators. The data and how the generators split the data amongst themselves are shown in Figure \ref{fig:toy_others} for MADGAN, MGAN, MEGAN. We color the different regions of $p(z)$ and the corresponding $p(x)$ so as to show which parts of $p(z)$ generate which part of $p(x)$; we also show in smaller plots below samples generated by individual generators similarly color coded. 

We see that because MADGAN and MGAN do not use any gating function, with these two, the output regions of the generators overlap with each other. Each color in $z$-space corresponds to eight different points in the $x$-space by the eight models. MEGAN does use a gating function but because the gating also uses extra $\nu_i(z)$, the output regions of generators still overlap with each other. Note that all three miss parts of the underlying distribution; MADGAN and MEGAN miss the top component, and MGAN miss the one at the bottom. 

The tree learned by our proposed hierarchical mixture of generators, HMoG, is shown in Figure \ref{fig:toy_hmog}; this is a tree of depth three that also has eight generators at its leaves. We calculate each decision node's responsibility by counting the (soft) gating values and an instance is drawn in the box of the expert having the highest responsibility. We see that the tree has learned a hierarchical soft clustering of the data with the leaves learning parts of $p(x)$ each corresponding to a part of $p(z)$. We see that this model covers the data completely and has not missed any of the components.

The results using a flat mixture of generators, MoG, is shown in Figure \ref{fig:toy_mog}. Because the combination is soft and only depends on the input $z$, here too each generator operates in a local region of $z$. This model too learns the distribution without dropping any modes of the data; note that here, we see that some generators do more of the work with some not used at all. This we believe is the advantage of a hierarchical model which dissects the problem into two at each level, easing the problem in a divide-and-conquer fashion. The hierarchical organization also lends to discovering structure in the data.

%

\subsection{Results on image data sets}
\label{subsec:results_image}

We test and compare our proposed mixture models HMoG and MOG with MADGAN, MGAN, and MEGAN, on five image data sets that are widely used in the GAN literature: MNIST \cite{lecun1998mnist}, FashionMNIST \cite{xiao2017fashion}, UTZap50K \cite{yu2014fine}, Oxford Flowers \cite{nilsback2008automated}, and CelebA \cite{liu2015deep}. We resize MNIST and FashionMNIST data set to $32 \times 32$ and the other data sets with more detail to $64 \times 64$. All image pixels are normalized to the range $[-1, 1]$.

It is known that using a convolutional architecture for tasks that involve images increases the performance dramatically, and we incorporate transposed convolutional (also known as deconvolutional or fractionally strided convolution) layers in each model. More specifically, we use the (transposed) convolutional part of DCGAN \cite{radford2015unsupervised} as the shared part of generators, denoted by $B_s$ above. Instead of generating samples directly in the data domain $x$, each model generates an abstract representation $h$ which is given to the shared block $B_s$ that produces the output $x$. For any data set, the same $B_s$ is used in all models.

All these variants combine multiple {\em local\/} models; we also define the fully connected (FC) model that uses a fully connected layer, which stands for the standard {\em distributed\/} alternative having one global generator, which we take as the baseline against which we compare all the localized variants.

In training HMoG, MoG, and the baseline model FC, Wasserstein loss with gradient penalty \cite{gulrajani2017improved} is used. For MADGAN and MGAN, we use the original likelihood-based loss; Wasserstein loss is not applicable with these since since they require $D$ to be a classifier. MEGAN can be used with either Wasserstein loss or the original loss; we use the original loss because it performed better in our preliminary experiments. For all methods, we used the Adam optimizer \cite{kingma2014adam} with amsgrad option \cite{reddi2019convergence}. The learning rate is set to $0.0001$ with beta values of Adam set to $(0.5, 0.999)$. The batch size is set to 128.

For evaluating the performance of the variants, we use the Fr\'echet Inception distance (FID) \cite{heusel2017gans} and the two-sample test (C2ST) \cite{lopez2016revisiting}, here, 5-nearest neighbor (5-NN) leave-one-out accuracy. Both FID and 5-NN accuracy are calculated with the activations before the softmax layer (2048-dim) of InceptionV3 \cite{szegedy2016rethinking}. Lower FID scores are better and 5-NN accuracies that are close to 50\% are better. All models are run five times with different random seeds, and we report the mean and standard deviations.

For flat models, we experiment with 4, 8, 16, and 32 generators, which for the hierarchical model translates to trees of depth 2, 3, 4, and 5. We also report the parameter count of each model; these do not include the shared deconvolution block used in all models.

\begin{figure*}[htbp]
\begin{center}
\subfloat[MNIST]{
	\includegraphics[width=0.99\columnwidth]{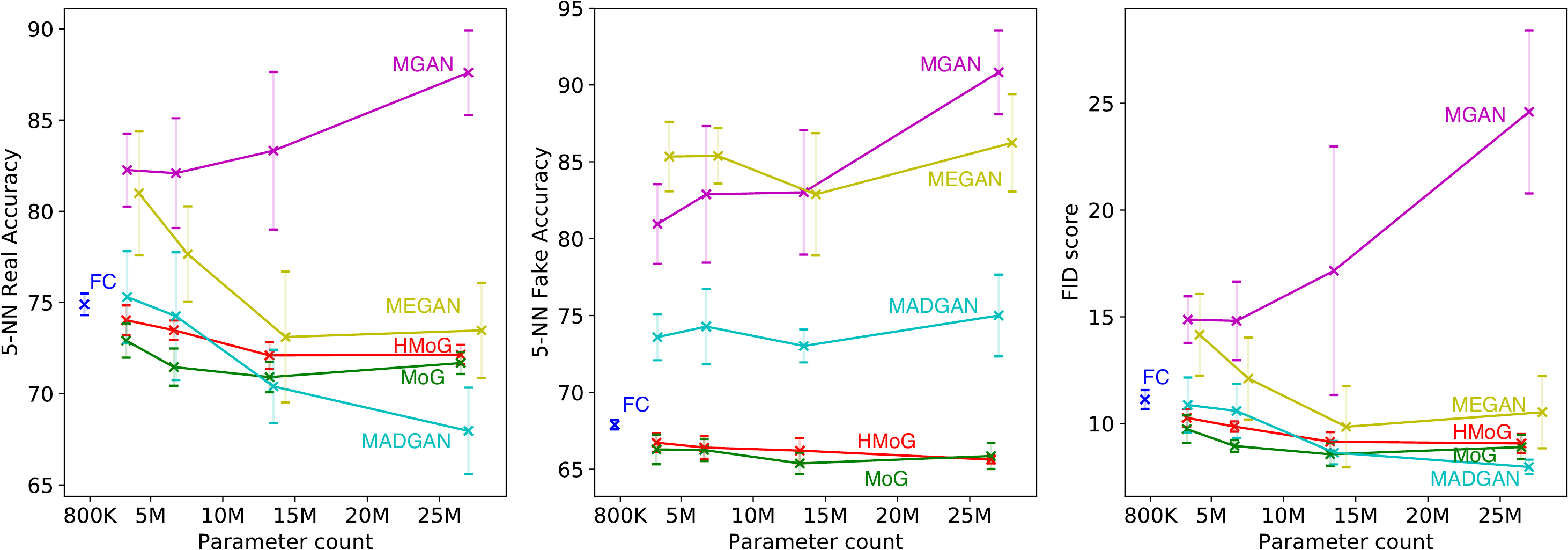}
	\label{fig:box_s:mnist}
}
\subfloat[FashionMNIST]{
	\includegraphics[width=0.99\columnwidth]{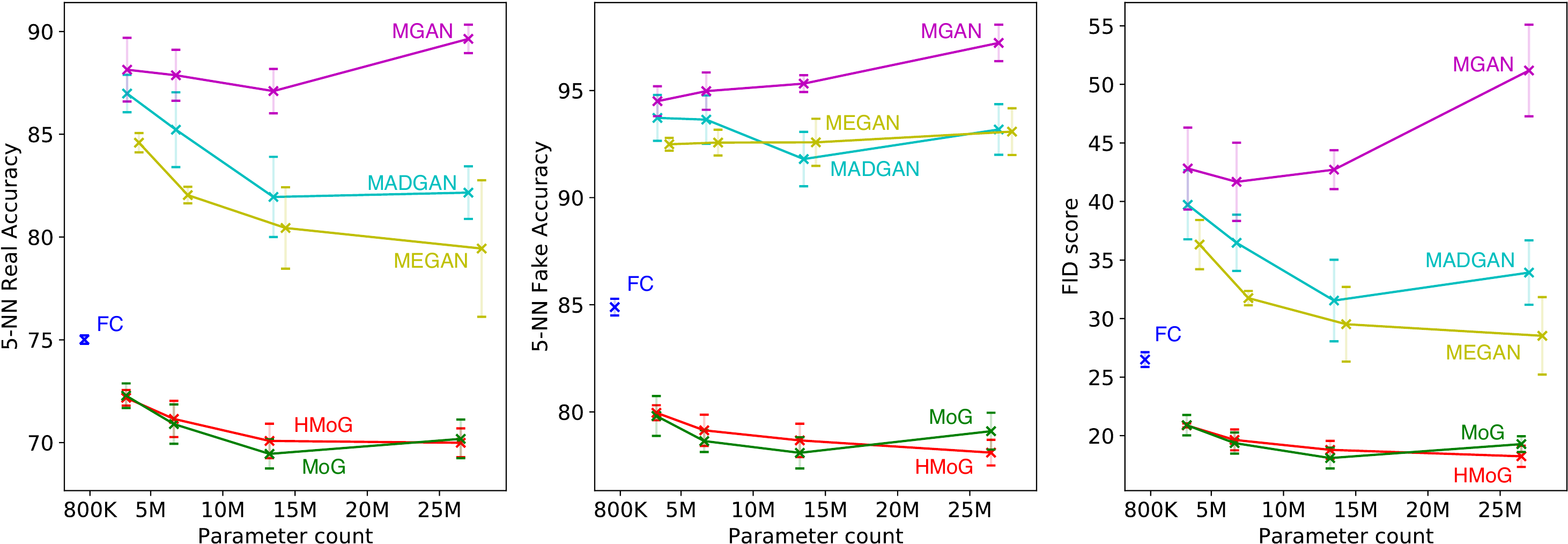}
	\label{fig:box_s:fashion}
}

\subfloat[UTZap50K]{
	\includegraphics[width=0.99\columnwidth]{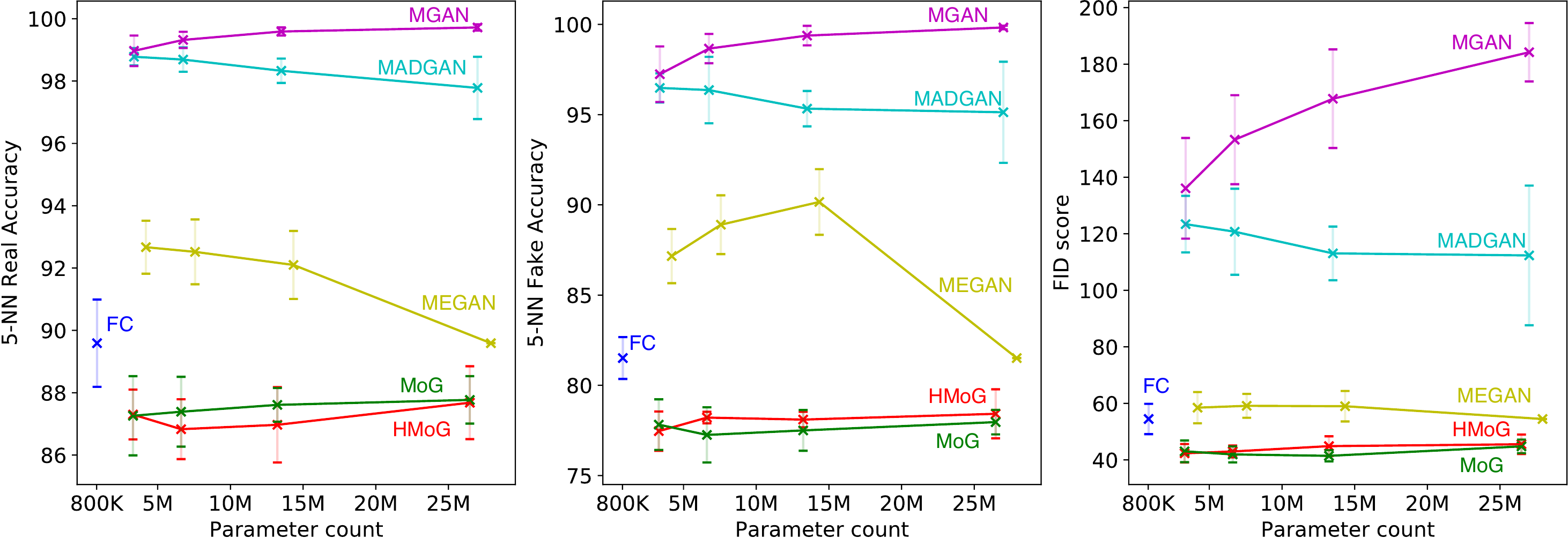}
	\label{fig:box_s:utzap50k}
}
\subfloat[Oxford Flowers]{
	\includegraphics[width=0.99\columnwidth]{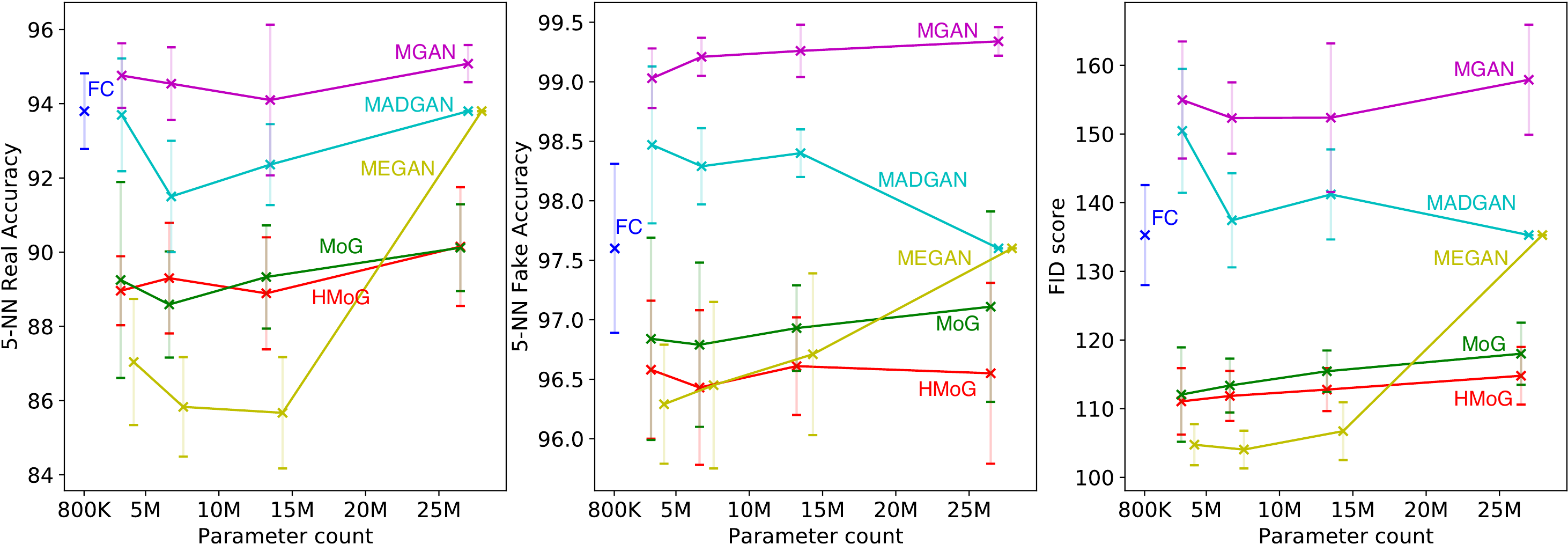}
	\label{fig:box_s:flowers}
}

\subfloat[CelebA]{
	\includegraphics[width=0.99\columnwidth]{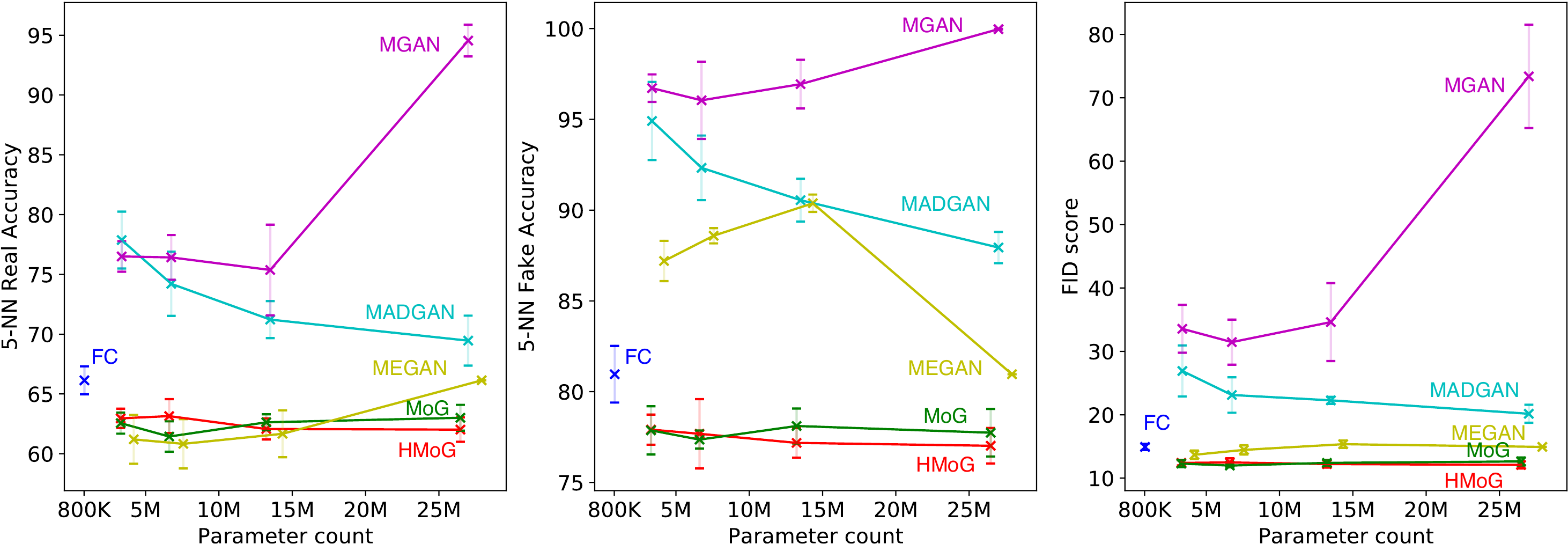}
	\label{fig:box_s:celeb}
}

\end{center}
\caption{FID scores and 5-NN accuracies of all tested models on the five data sets for different number of generators. These are average and one standard deviation error bars of five runs. The $x$-axis is the parameter count (not including $B_s$ used in all) and the $y$-axis represents 5-NN real accuracy, 5-NN fake accuracy, or the FID score, respectively. Lower FID and 5-NN scores closer to 50\% are better.}
\label{fig:results}
\end{figure*} 

Our experimental results on the five data sets are shown in Figure \ref{fig:results}. We see that in terms of FID score, both of our proposed MoG and HMoG outperform other approaches. We also see that MADGAN and MGAN perform worse than the baseline FC; only on MNIST, MADGAN performs better than the baseline. This might suggest that forcing discriminator to classify generators may not always work, which is the idea behind MADGAN and MGAN. On the other hand, MEGAN seems to perform on par with the baseline, sometimes even better. Note that unlike MADGAN and MGAN, MEGAN uses a gating function to select among its generators. This hints at the importance of training different generators in different input regions and combining them based on the input, instead of relying on the discriminator to force multiple generators to different modes.

If we compare our mixture of experts formulation (MoG) with MEGAN, we see that our model gets better results in terms of FID scores and 5-NN accuracies. As opposed to MEGAN, our mixture of generators is a soft cooperative one. The input to the gating model is only the latent $z$, which also reduces the number of parameters significantly.

\begin{figure*}[htbp]
\begin{center}
\subfloat{
    \includegraphics[width=0.99\columnwidth]{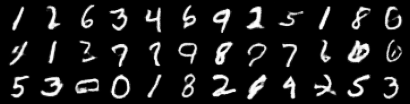}
}
\subfloat{
    \includegraphics[width=0.99\columnwidth]{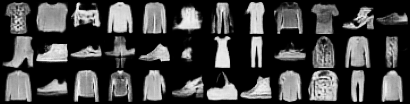}
}

\subfloat{
    \includegraphics[width=0.99\columnwidth]{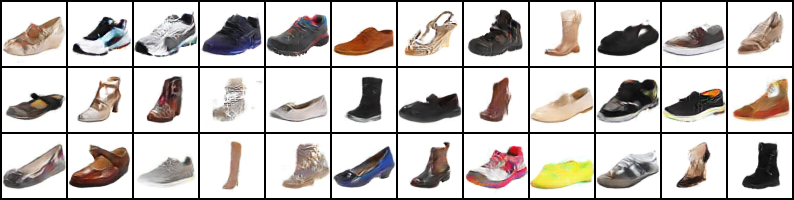}
}
\subfloat{
    \includegraphics[width=0.99\columnwidth]{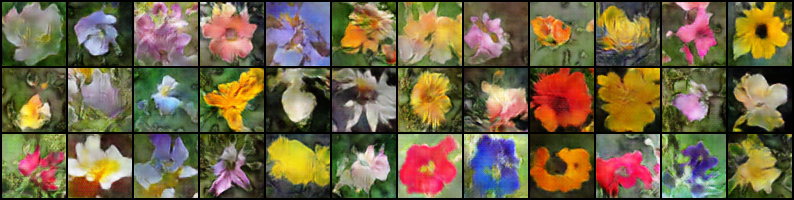}
}

\subfloat{
    \includegraphics[width=0.99\columnwidth]{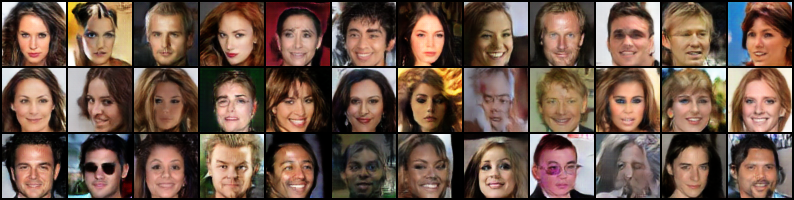}
}
\end{center}
\caption{Samples generated by the hierarchical mixture of generators (HMoG) with depth four (and 16 generators) on the five data sets.}
\label{fig:samples_hmog}
\end{figure*}

Some samples generated from HMoG with depth four are shown in Figure \ref{fig:samples_hmog}. For the sampling procedure, we randomly draw $\{z_i\}$ and disregard the least likely $25\%$ percent to get rid of possible outliers \cite{brock2018large}. A visual inspection of these also show that HMoG is able to generate realistic and diverse samples on all data sets.

\subsection{Interpretability}
\label{subsec:interpretation}

\begin{figure}[htbp]
\begin{center}
\subfloat[MoG]{
	\includegraphics[width=0.47\columnwidth]{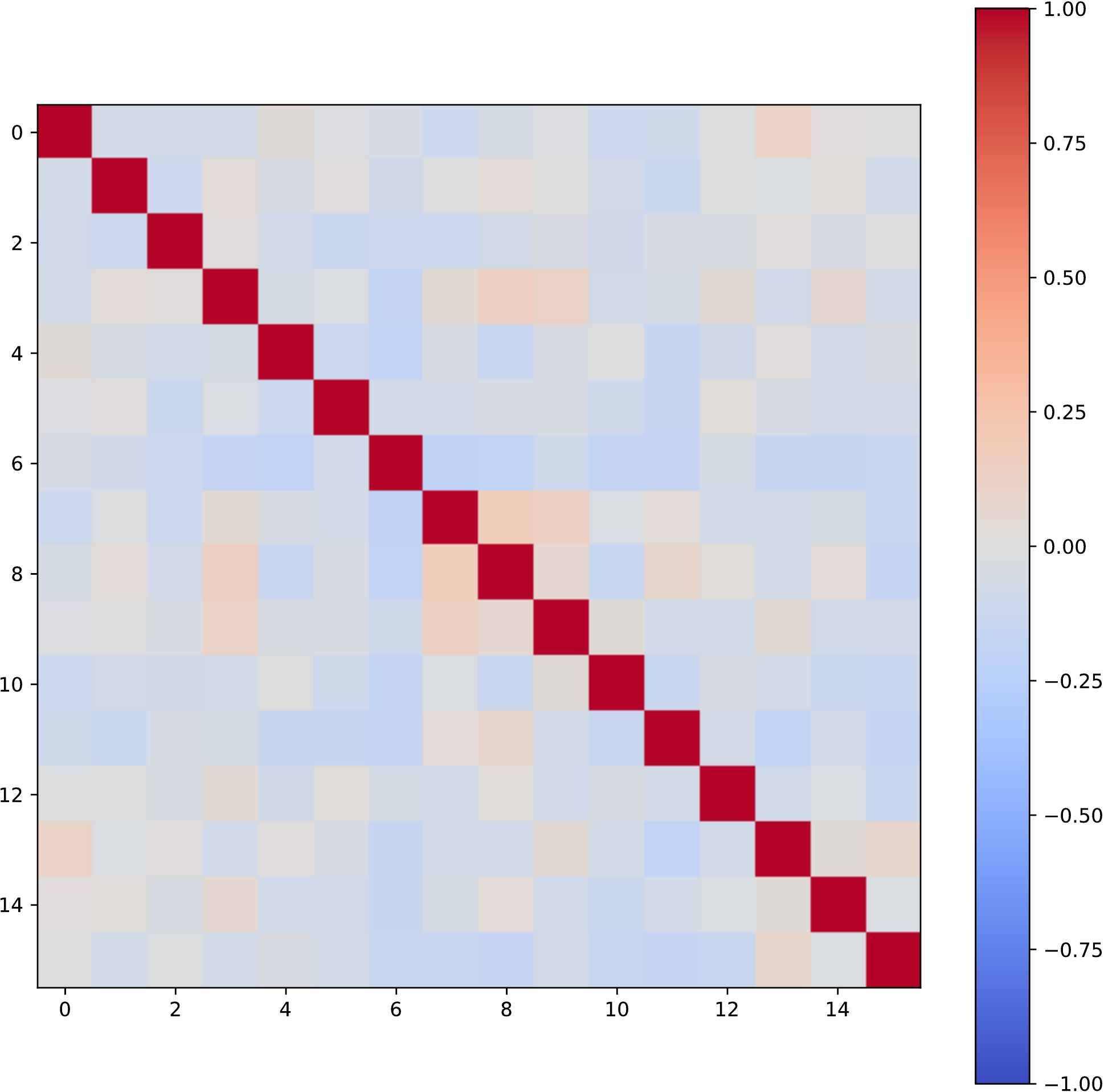}
	\label{fig:corr:mog}
}
\subfloat[HMoG]{
	\includegraphics[width=0.47\columnwidth]{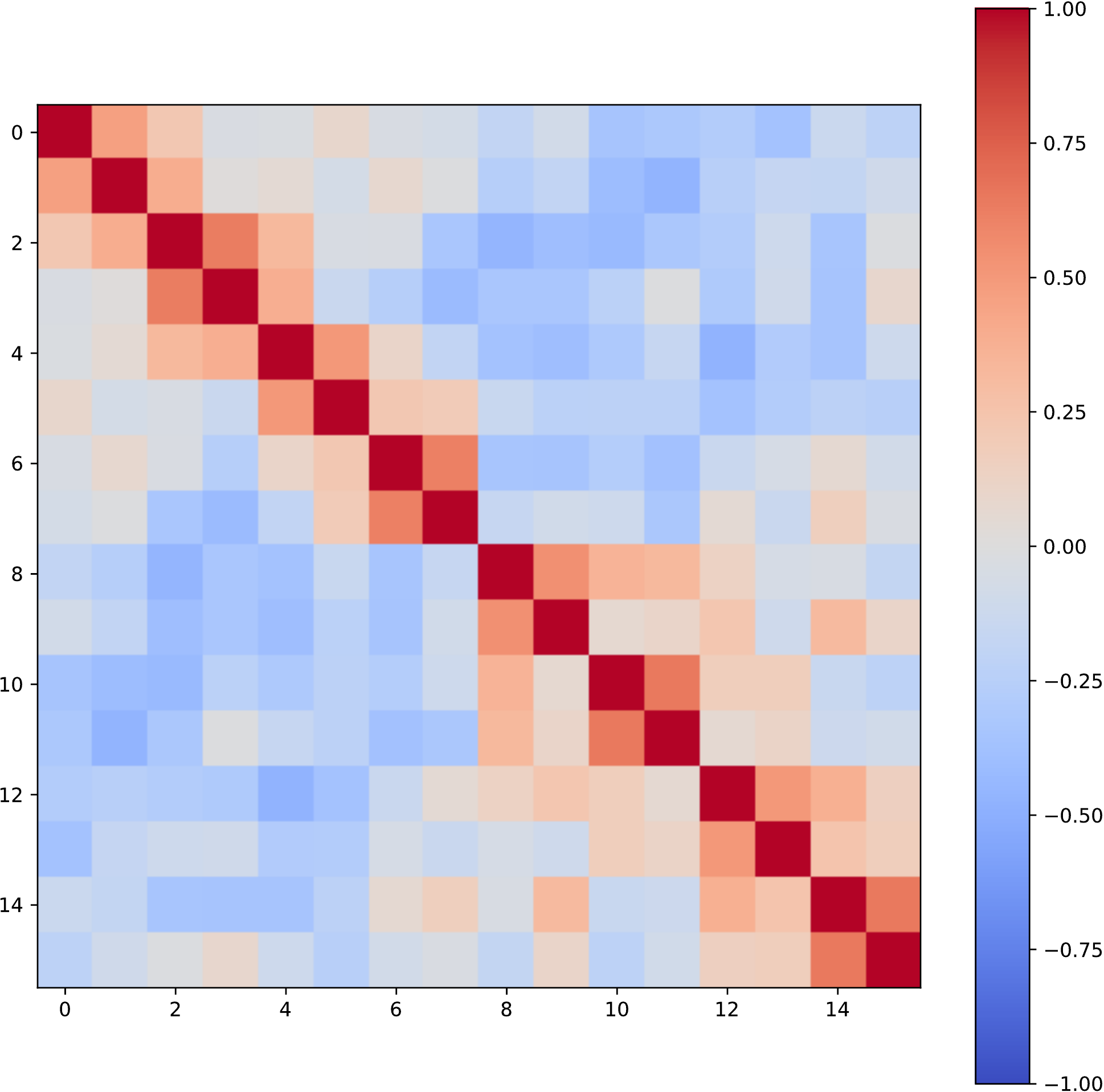}
	\label{fig:corr:hmog}
}
\end{center}
\caption{Correlation matrices of gating values of generators for MoG and HMoG with 16 generators. With MoG, there is no apparent correlation; with HMoG, we see that generators that are closer in the tree (in the same subtree) are used together, which imply a semantic correlation.}
\label{fig:correlations}
\end{figure}

Because both MoG and HMoG use a soft combination, we can check whether there is any correlation between the outputs of the local generators. For the flat MoG, the probability that a local model is used is given by the softmax gating; for the HMoG, it is the product of all the (binary) gatings on the path to the root. We calculate the correlation between these probabilities for pairs of local models for the case of 16 generators (or a tree of depth 4) on the CelebA data set. The $16 \times 16$ correlation matrices for both are shown color coded in Figure \ref{fig:correlations}. 

We see that with the flat MoG, correlations are randomly scattered. In HMoG however, we see that the correlations are gathered around the diagonal; we can see spectral squares of sizes $2 \times 2$ and $4 \times 4$ corresponding to subtrees, which is an implication that generators that have the same ancestor on the second or the third level of the tree are frequently used together indicating that they learn semantically correlated samples.

\begin{figure}[htbp]
    \begin{center}
    \includegraphics[width=1\columnwidth]{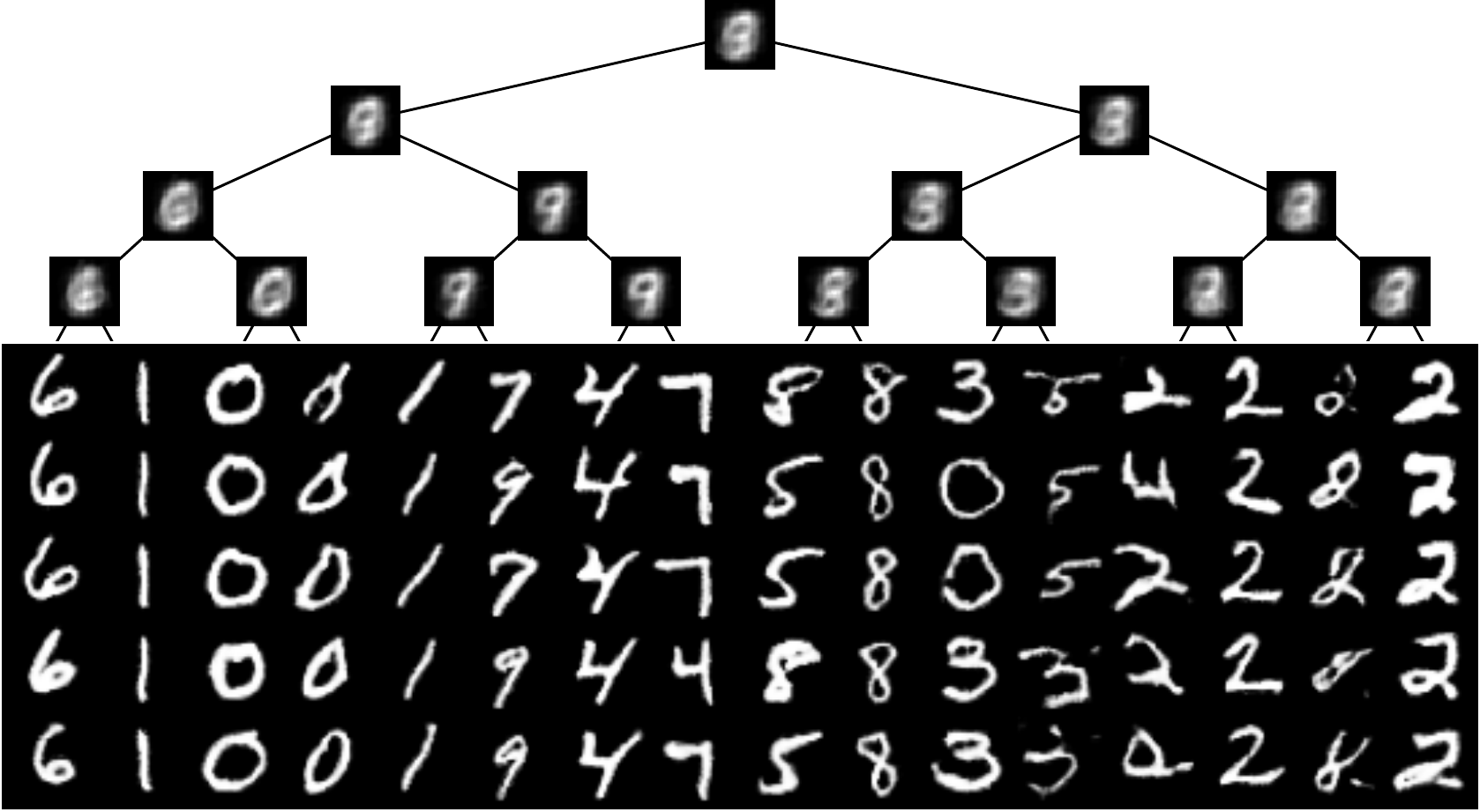}
    \end{center}
    \caption{On MNIST using HMoG, the average response of each internal node in the tree hierarchy is shown. For each leaf, five random samples are shown that have the highest probability of being generated in there.}
    \label{fig:tree:mnist}
\end{figure} 

\begin{figure*}[htbp]
    \begin{center}
    \includegraphics[width=1\textwidth]{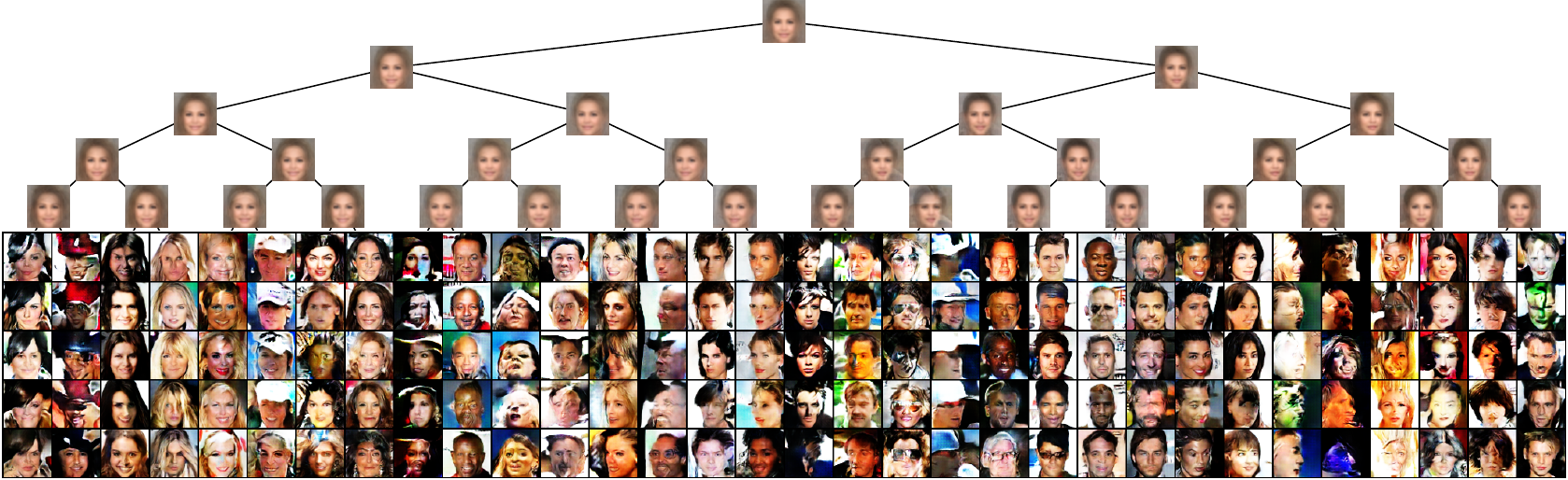}
    \end{center}
    \caption{On Celeb using HMoG, the average response of each internal node in the tree hierarchy is shown. For each leaf, five samples are shown that have the highest probability of being generated in there.}
    \label{fig:tree:celeb}
\end{figure*} 

In Figures \ref{fig:tree:mnist} and \ref{fig:tree:celeb}, the average responses of decision nodes in the tree are visualized by taking the weighted average of the generated samples on MNIST and CelebA respectively. For a given node, weights are found by multiplying the gating probabilities along the path to the node. At the bottom of the tree under each leaf, we show five random samples generated from the corresponding generator. To find these, we sample random 10,000 $z$ vectors and select the top five most likely for each generator. Here, the most likely point for a generator is the point which maximizes the probability that the corresponding leaf is chosen. We see the data set mean at the top root, and as we go down the tree the blurriness decreases and each node becomes more specialized to a specific region of $p(x)$. We see in Figure \ref{fig:tree:mnist} that digits that are similar in shape are generated by leaves that are nearby in the tree. For CelebA too, as we see in Figure \ref{fig:tree:celeb}, we see that the examples are distributed over the leaves in terms of similarity in orientation, color, or background.

We believe that this interpretability is the advantage of the HMoG model over the MoG model, as well as other approaches that train a flat set of generators. As in soft hierarchical clustering, the division at each level, which may be interpreted as an architectural inductive bias, lets us view the data in different levels of granularity and understand the decisive features of the data through a divide-and-conquer type of approach.

\section{Conclusions}
\label{sec:conclusion}

We propose the hierarchical mixture of generators, HMoG, and a special case, MoG, which is a flat mixture of generators. There are GAN variants in the literature that also combine multiple generators but they are limited in the way they force the generators to different modes. Our formulation is the first to our knowledge that learns a cooperative mixture of generators, either organized in a flat manner or hierarchically.

An important advantage of the hierarchical model is its interpretability. Since it is a tree architecture, we can make a post-hoc analysis of the learned tree to gain insight about the data. At each level of the tree, nodes can be seen as clusters, or modes, in different levels of granularity, where as we go down the tree, clusters get more local. At the same time, splits are soft and what the tree learns is a hierarchical soft clustering of the data. In the generative setting that we have here, the leaves are generators each responsible from generating one local cluster, 

Our experimental results on five data sets show that the proposed models can generate samples that are realistic and diverse. Our proposed models have better FID score and 5-NN accuracy with lower variance when compared with other methods that incorporate multiple generators as well as the fully-connected standard GAN implementation.

\section*{Acknowledgements}
This work is partially supported by Bo\u{g}azi\c{c}i University Research Funds with Grant Number 18A01P7. The numerical calculations reported in this work were partially performed at TUBITAK ULAKBIM, High Performance and Grid Computing Center (TRUBA resources).

\bibliographystyle{plain} 
\bibliography{references}



\end{document}